# AUTOMATIC ANALYSIS OF EEGS USING
# BIG DATA AND HYBRID DEEP LEARNING ARCHITECTURES

Meysam Golmohammadi[a], Amir Hossein Harati Nejad Torbati[b],
Silvia Lopez de Diego[a], Iyad Obeid[a], and Joseph Picone[a]

[a] The Neural Engineering Data Consortium, Temple University, Philadelphia, Pennsylvania, USA
[b] Jibo, Inc., Redwood City, California, USA

*Abstract*

*Objective:* A clinical decision support tool that automatically interprets EEGs can reduce time to diagnosis and enhance real-time applications such as ICU monitoring. Clinicians have indicated that a sensitivity of 95% with a specificity below 5% was the minimum requirement for clinical acceptance. We propose a high-performance classification system based on principles of big data and machine learning.

*Methods:* A hybrid machine learning system that uses hidden Markov models (HMM) for sequential decoding and deep learning networks for postprocessing is proposed. These algorithms were trained and evaluated using the TUH EEG Corpus, which is the world's largest publicly available database of clinical EEG data.

*Results:* Our approach delivers a sensitivity above 90% while maintaining a specificity below 5%. This system detects three events of clinical interest: (1) spike and/or sharp waves, (2) periodic lateralized epileptiform discharges, (3) generalized periodic epileptiform discharges. It also detects three events used to model background noise: (1) artifacts, (2) eye movement (3) background.

*Conclusions:* A hybrid HMM/deep learning system can deliver a low false alarm rate on EEG event detection, making automated analysis a viable option for clinicians.

*Significance:* The TUH EEG Corpus enables application of highly data consumptive machine learning algorithms to EEG analysis. Performance is approaching clinical acceptance for real-time applications.

## Highlights:

- A hybrid machine learning system based on hidden Markov models and deep learning is proposed for automatic interpretation of EEGs.

- The results are reported on TUH EEG Corpus, which is the world's largest publicly available database of EEG recordings.

- Using big data and deep learning, performance is approaching that required for clinical acceptance.

## 1.   INTRODUCTION

Electroencephalograms (EEGs) are used in a wide range of clinical settings to record electrical activity along the scalp. EEGs are the primary means by which physicians diagnose brain-related illnesses such as epilepsy and seizures (Yamada et al., 2017). However, analysis of EEG signals requires a highly-trained neurophysiologist. Manual analysis is time-consuming and expensive since identifying rare clinical events requires analysis of long data streams. Automatic analysis of EEG scans reduces time to diagnosis, reduces error and enhances real-time applications by flagging sections of the signal that need further review by a clinician. Many methods have been developed over the years (Ney et al., 2016) including time–frequency analysis methods (Gotman, 1999; Sartoretto & Ermani, 1999; Osorio et al., 1998), nonlinear techniques (Schad et al., 2008; Stam, 2005; Schindler et al., 2001) and expert systems that attempt to mimic a human observer (Deburchgraeve et al., 2008; Khamis et al., 2009). Despite much progress and research, current EEG analysis methodologies are far from perfect with many being considered impractical due to high false detection rates (Hopfengärtner et al., 2007; Varsavsky & Mareels, 2006).

Machine learning has made tremendous progress over the past three decades due to rapid advances in low-cost highly-parallel computational infrastructure, powerful machine learning algorithms, and, most importantly, big data. Although contemporary approaches for automatic interpretation of EEGs have employed more modern machine learning approaches such as neural networks (Ramgopal et al., 2014) and support vector machines (Alotaiby et al., 2014), state of the art machine learning algorithms that employ high dimensional models have not previously been utilized in EEG analysis because there has been a lack of large databases that incorporate sufficient real world variability to adequately train these systems. In fact, what has been lacking in many bioengineering fields including automatic interpretation of EEGs are the big data resources required to support the application of advanced machine learning approaches. A significant big data resource, known as the TUH EEG Corpus (Obeid & Picone, 2016), has recently become available creating a unique opportunity to evaluate high performance deep learning models that require large amounts of training data. This database includes detailed physician reports and patient medical histories, which are critical to the application of deep learning. But, transforming physicians' reports into a deep learning paradigm is proving to be challenging because the mapping of reports to underlying EEG events is nontrivial. Our experiments suggest that a hybrid approach based on hidden Markov models and deep learning can approach clinically acceptable levels of performance.

## 2.   METHOD

An overview of our proposed system is shown in Fig. 1. An $N$-channel EEG is transformed into $N$ independent feature streams using a standard sliding window based approach. A sequential modeler analyzes each channel and produces feature hypotheses. Three passes of postprocessing are performed to produce the final output. In this section, we discuss the various components of this system, including development of the statistical models using a supervised training approach. We begin with a discussion of the data used to train and evaluate the system.

### 2.1.   Data: The TUH EEG Corpus

Our system was developed using the TUH EEG Corpus (TUH-EEG) (Obeid & Picone, 2016), which is the world's largest publicly available database of clinical EEG data. It contains over 30,000 sessions from over 16,000 patients (over 30 years of signal data in total). It is an ongoing data collection effort. The most recent release, v1.0.0, includes data from 2002 – 2015. This EEG data was collected at the Department of Neurology at Temple University Hospital. It is entirely composed of clinical data with all the real-world artifacts one would expect to see in clinical recordings (e.g., eye blinking and head movements). Each of the sessions contains at least one EDF file and one physician report. These reports are generated by a board-certified neurologist and are the official hospital record. These reports are comprised of unstructured text

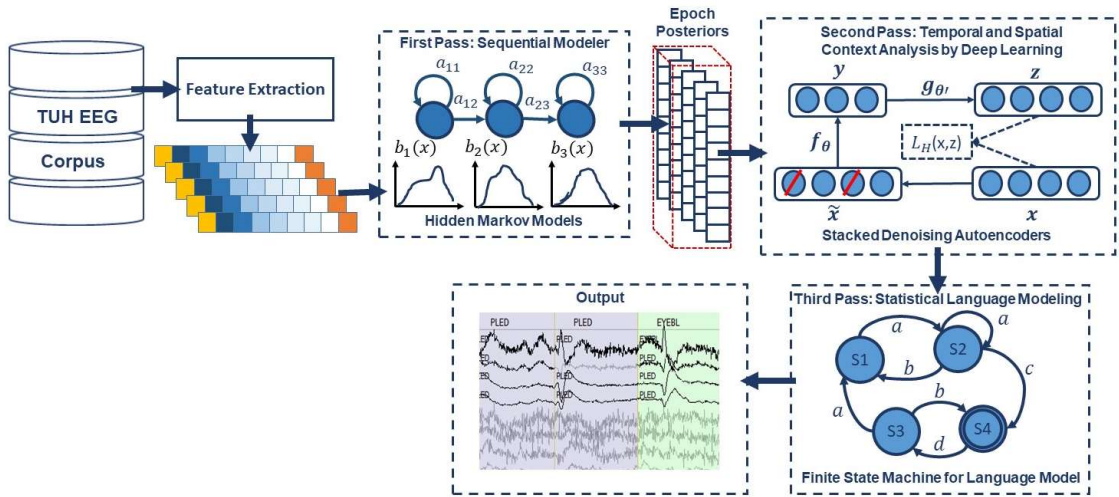

Fig. 1. A three-pass architecture for automatic interpretation of EEGs that integrates hidden Markov models for sequential decoding of EEG events with deep learning for decision-making based on temporal and spatial context

that describes the patient, relevant history, medications, and clinical impression. The corpus is publicly available from the Neural Engineering Data Consortium (*www.nedcdata.org*).

EEG signals in TUH-EEG were recorded using several generations of Natus Medical Incorporated's Nicolet™ EEG recording technology. The raw signals consist of multichannel recordings in which the number of channels varies between 20 and 128 channels (Harati et al., 2014). A 16-bit A/D converter was used to digitize the data. The sample frequency varies from 250 Hz to 1024 Hz. In our study, we have resampled all the data to a common sample frequency of 250 Hz. The Natus system stores the data in a proprietary format that has been exported to EDF with the use of NicVue v5.71.4.2530. The original EEG records are split into multiple EDF files depending on how the session was annotated by the attending technician. Some statistics about the corpus are shown in Fig. 2.

A portion of TUH-EEG was annotated manually during a study conducted with Temple University Hospital neurologists (Harati et al., 2014). These annotations comprise six types of events. The first three events are of clinical interest: (1) spike and/or sharp waves (SPSW), (2) periodic lateralized epileptiform discharges (PLED), and (3) generalized periodic epileptiform discharges (GPED). SPSW events are epileptiform transients that are typically observed in patients with epilepsy. PLED events are indicative of EEG abnormalities and often manifest themselves with repetitive spike or sharp wave discharges that can be focal or lateralized over one hemisphere. These signals display quasi-periodic behavior. GPED events are similar to PLEDs, and manifest themselves as periodic short-interval diffuse discharges, periodic long-interval diffuse discharges and suppression-burst patterns according to the interval between the discharges. Triphasic waves, which manifest themselves as diffuse and bilaterally synchronous spikes with bifrontal predominance, typically at a rate of 1-2 Hz, are also included in this class.

The remaining three events were used by our machine learning technology to model background noise: (1) eye movement (EYEM), (2) artifacts (ARTF), and (3) background (BCKG). EYEM events are spike-like signals that occur during patient eye movement. These are quite common in clinical data, so for reasons explained later, we devoted a specific class to these events. ARTF events other than EYEM are recorded electrical activity that is not of cerebral origin, such as those due to the equipment, patient behavior or the environment. These are common events that can often be confused with SPSW, and hence, it is important such things be included in the database. BCKG is used to annotate all other portions of the signal.

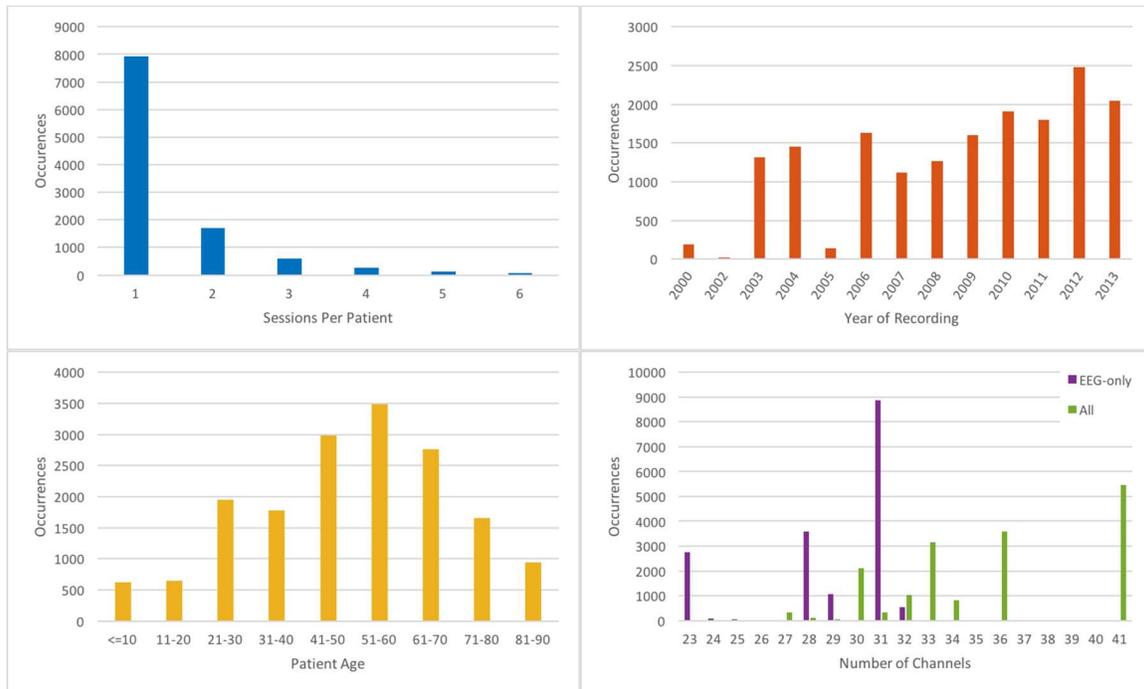

Fig. 2. Some relevant statistics demonstrating the variety of data in TUH-EEG

There are over 10 different electrode configurations and over 40 channel configurations represented in the corpus. This poses a serious challenge for machine learning systems since for a system to be practical it must be able to adapt to the specific type of EEG being administered. However, for this initial study, we focused on a subset of the data in which signals were recorded using the Averaged Reference (AR) electrode configuration (Lopez et al., 2016). The next step in the data pipeline is to convert the data to a feature representation.

## 2.2.   Preprocessing: Feature Extraction

The first step in EEG processing in Fig. 1 consists of converting the signal to a sequence of feature vectors (Picone, 1990). Feature extraction for automatic classification of EEG signals typically relies on time frequency representations of the signal (Thodoroff et al., 2016; Mirowski et al., 2009). While techniques such as cepstral-based filter banks or wavelets are popular analysis techniques in many signal processing applications including EEG classification (Subasi et al., 2007; Jahankhani et al., 2006), our system uses a standard cepstral coefficient-based feature extraction approach based on Linear Frequency Cepstral Coefficients (LFCCs) that has been popular in applications such as speech recognition (Picone, 1990). We use a linear frequency scale for EEGs because there is little evidence that a nonlinear scale is relevant to this problem, and in practice performance was slightly better for a linear scale (Harati et al., 2015). Recent experiments with different types of features (Da Rocha Garrit et al., 2015) or with using sampled data directly (Xiong et al., 2017) have not shown a significant improvement in performance by eliminating the feature extraction process and using sampled data directly.

It is common in the LFCC approach to compute cepstral coefficients by computing a high resolution fast Fourier Transform, downsampling this representation using an oversampling approach based on a set of overlapping bandpass filters, and transforming the output into the cepstral domain using a discrete cosine transform (Picone, 1990). In this study, the zeroth-order cepstral term is discarded and replaced with a frequency domain energy term by computing the sum of squares of the oversampled filter bank outputs after they are downsampled:

$$E_f = \log\left(\sum_{k=0}^{N-1}|X(k)|^2\right) . \tag{1}$$

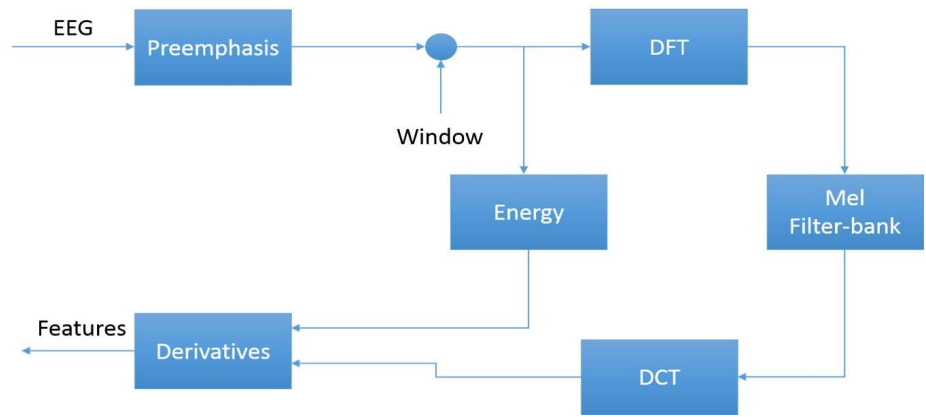

Fig. 3. An overview of the feature extraction algorithm

In order to improve differentiation between transient pulse-like events and stationary background noise, we have introduced a differential energy term that attempts to model the long-term change in energy. This term examines energy over a range of M frames centered about the current frame, and computes the difference between the maximum and minimum over this interval:

$$E_d = \max_m\left(E_f(m)\right) - \min_m\left(E_f(m)\right) . \tag{2}$$

We typically use a 0.9 sec window for this calculation. This simple feature has proven to be surprisingly effective (Harati et al., 2015).

The final step to note in our feature extraction process is a familiar method for computing derivatives of features using a regression approach:

$$d_t = \frac{\sum_{n=1}^{N} n(c_{t+n} - c_{t-n})}{2\sum_{n=1}^{N} n^2}, \tag{3}$$

where $d_t$ is a delta coefficient, from frame $t$ computed in terms of the static coefficients $c_{t+n}$ to $c_{t-n}$. A typical value for $N$ is $9$ (corresponding to $0.9$ secs) for the first derivative in EEG processing, and $3$ for the second derivative. The introduction of derivatives helps the system discriminate between steady-state behavior, such as a PLED, and impulsive or nonstationary signals such as spikes and eye movements.

These features, which are often called deltas because they measure the change in the features over times, are one of the most well-known features in speech recognition. We typically use Eq. (3) to compute the derivatives of the features and then apply this approach again to those derivatives to obtain an estimate of the second derivatives of the features, generating what are often called delta-deltas. This process triples the size of the feature vector (adding deltas and delta-deltas), but is well-known to deliver small but measurable improvements in performance.

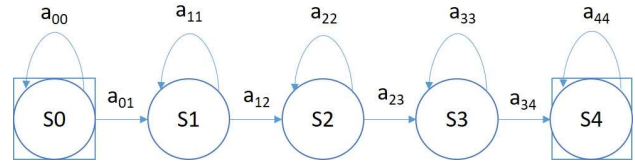

Fig. 4. A left-to-right HMM is used for sequential decoding in the first pass of processing.

In this work, through experiments designed to optimize feature extraction, we found best performance can be achieved using a feature vector length of *26* comprising *7* cepstral coefficients, *1* frequency energy, *1* differential energy, *9* deltas for all and *8* delta-deltas for just cepstral terms plus frequency energy (Harati et al., 2015). A block diagram of the feature extraction process used in this work for automatic classification of EEG signals is presented in **Error! Reference source not found.**.

### 2.3. First Pass: Sequential Decoding Using Hidden Markov Models

Hidden Markov Models (HMMs) are among the most powerful statistical modeling tools available today for signals that have both a time and frequency domain component (Juang and Rabiner, 1991; Picone, 1990). HMMs are a class of doubly stochastic processes in which discrete state sequences are modeled as Markov chains. HMMs have been used extensively in speech recognition where a speech signal can be decomposed into an energy and frequency profile in which particular events in the frequency domain can be used to identify the sound spoken. The challenge of interpreting and finding patterns in EEG signal data is very similar to that of speech related projects with a measure of specialization.

Like speech recognition systems, we assume that the EEG signal is a realization of some message encoded as a sequence of one or more symbols. We model an EEG as a sequence of one of six symbols: SPSW, PLED, GPED, EYEM, ARTF and BCKG. Let each event be represented by a sequence of feature vectors or observations O, defined as:

$$O = o_1, o_2, \ldots, o_T \ . \tag{4}$$

Here $o_T$ is the feature vector observed at time t. Then considering $S_i$ is the $i^{th}$ event in our dictionary, the isolated event recognition problem can be regarded as finding the most probable event which for a given set of prior probabilities, $P(S_i)$, depend only on the likelihood $P(O|S_i)$. We train one HMM model for each event using manually annotated data.

A simple left-to-right GMM-HMM, illustrated in Fig. 4, was used for sequential decoding of EEG signals. A GMM-HMM is characterized by $N$ number of states, $L$-component Gaussian mixture model, the transition probability $a_{ij}$ from state $i$ to $j$ and the output probability $b_{ij}(o)$ for symbol $o$ in the transition process. Considering $\alpha(i,t)$ as the forward probability where ($i = 1, 2, \ldots, N; t = 1, 2, \ldots, T$) , $\beta(j,t)$ as the backward probability where ($j = 1, 2, \ldots, N; t = T-1, \ldots, 0$), and $P(O|M)$ as the probability that model $M$ generates symbol series $O$, the probability that a transition from state $i$ to state $j$ happens at time $t$ can be defined as:

$$\gamma_i(i,j) = \frac{\alpha(i,t-1)a_{ij}b_{ij}(O_t, \mu_{ij}, \Sigma_{ij})\beta(j,t)}{P(O|M)} \ . \tag{5}$$

The reestimation formulae for the transition probabilities are:

$$a_{ij} = \frac{\sum_t \gamma_i(i,j)}{\sum_t \sum_j \gamma_i(i,j)} \ . \tag{6}$$

If the output vector, $O_t$, follows an *n*-dimensional normal distribution, the output density function is as follows:

$$b_{ij}(o_t, \mu_{ij}, \Sigma_{ij}) = \frac{exp\{-(o_t-\mu_{ij})^t \Sigma_{ij}^{-1}(o_t-\mu_{ij})/2\}}{(2\pi)^{n/2}|\Sigma_{ij}|^{1/2}}, \tag{7}$$

where $\mu_{ij}$ is the mean and $\Sigma_{ij}$ is the covariance matrix. The mean and covariance for each Gaussian mixture component can be estimated by:

$$\mu_{ij} = \frac{\sum_t \gamma_i(i,j)o_t}{\sum_t \gamma_i(i,j)}, \tag{8}$$

$$\Sigma_{ij} = \frac{\sum_t \gamma_i(i,j)(o_t-\mu_{ij})(o_t-\mu_{ij})^t}{\sum_t \gamma_i(i,j)}. \tag{9}$$

In the first pass of signal modeling shown in Fig. 1, we divide each channel of the EEG signal into epochs. Then each epoch is represented by a sequence of frames where each frame is represented by a feature vector. During training, we estimate the parameters of the $K$ models ($a_{ij}, b_{ij}, \mu_{ij}$ and $\Sigma_{ij}$) from the training dataset by iterating over all epochs using Eqs. (5-9). To determine these parameters in an iterative fashion, it is first necessary to initialize them with a carefully chosen value (Picone, 1990). Once this is done, more accurate parameters, in the maximum likelihood sense, can be found by applying the so-called Baum-Welch reestimation algorithm (Picone, 1990). Decoding is typically performed using the Viterbi algorithm (Alphonso et al., 2004). Then using one HMM model per label, we generate one posterior probability for each model and we select the label that corresponds to the highest probability.

## 2.4. Second Pass: Temporal and Spatial Context Analysis Based on Deep Learning

The goal of the second pass of processing in Fig. 1 is to integrate spatial and temporal context to improve decision-making. Therefore, the output of the first pass of processing, which is a vector of six posterior probabilities for every epoch of each channel, postprocessed by a deep learning system. Deep learning technology automatically self-organizes knowledge in a data-driven manner and learns to emulate a physician's decision-making process.

Deep Learning allows computational models that are composed of multiple processing layers to learn representations of data with multiple levels of abstraction. These methods have dramatically improved the state-of-the-art in speech recognition and many other domains in recent years (LeCun, et al., 2015). Deep learning discovers intricate structure in large data sets by using the backpropagation algorithm to indicate how a machine should change its internal parameters that are used to compute the representation in each layer from the representation in the previous layer.

In the second pass of processing, we are using a specific type of deep leaning network known as a Stacked denoising Autoencoders (SdA) (Vincent et al., 2010). SdAs have proven to perform well for applications where we need to emulate human knowledge (Bengio et al., 2007). Since interrater agreement for annotation of seizures tends to be relatively low and somewhat ambiguous, we need a deep learning structure that can deal with noisy inputs. From a structural point of view, SdAs are a form of stacking denoising autoencoders that form a deep network by using the latent representation of the denoising autoencoder found in the layer below as the input to the current layer.

Denoising Autoencoders are themselves an extension of a classical autoencoder (Vincent et al., 2008). An autoencoder takes an input vector $x \in [0,1]^d$, and first maps it to a hidden representation $y \in [0,1]^{d'}$ through a deterministic mapping:

$$y = f_\theta(x) = s(Wx + b), \tag{10}$$

where $W$ is a $d' \times d$ weight matrix, $b$ is a bias vector, $s$ is a nonlinearity such as Sigmoid function and $\theta = \{W, b\}$. The latent representation $y$, or code, is then mapped back, with a decoder, into a reconstruction $z$ of the same shape as $x$:

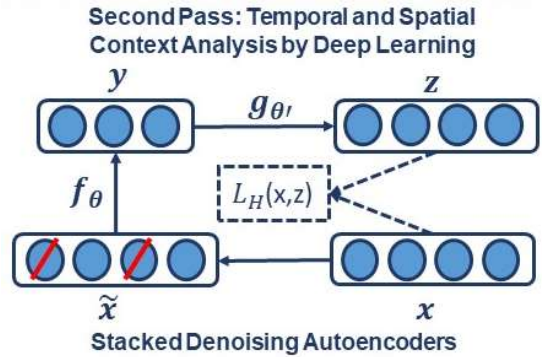

Fig. 5. In a stacked denoising autoencoder the input, $x$, is corrupted to $\tilde{x}$. The autoencoder then maps it to $y$ and attempts to reconstruct $x$.

$$z = g_{\theta'}(y) = s(W'y + b') . \qquad (11)$$

The weight matrix $W'$ of the reverse mapping may optionally be constrained by $W' = W^T$, in which case the autoencoder is said to have tied weights. The parameters of this model are optimized to minimize the average reconstruction error using a loss function, $L$, such as reconstruction cross-entropy:

$$\theta^*, \theta'^* = \underset{\theta, \theta'}{arg\ min}\ \frac{1}{n}\sum_{i=1}^{n} L\big(x^{(i)}, g_{\theta'}\big(f_\theta(x^{(i)})\big)\big) . \qquad (12)$$

To implement a denoising autoencoder, we train an autoencoder to reconstruct a clean "repaired" input from a corrupted, partially destroyed one. This is done by first corrupting the initial input $x$ to get a partially destroyed version $\tilde{x}$ by means of a stochastic mapping $\tilde{x} = q_D(\tilde{x}|x)$. Then this corrupted input is mapped, similar to a basic autoencoder, to a hidden representation $y = f_\theta(\tilde{x}) = s(W\tilde{x} + b)$ from which we reconstruct $z = g_{\theta'}(y) = s(W'x + b')$. The schematic representation of the process is presented in Fig. 5. As before, the parameters are trained to minimize the average reconstruction error over a training set, making $z$ as close as possible to the uncorrupted input $x$. But the key difference is that $z$ is now a deterministic function of $\tilde{x}$ rather than $x$ and thus the result of a stochastic mapping of $x$.

The application of deep learning networks like SdAs generally involves three steps: design, training and implementation. In the design step, the number of inputs and outputs, the number of layers, and the function of nodes are defined. During training, the weights of nodes are determined through a deep learning process. In the last step, the statistical model is implemented using the fixed parameters of the network determined during training. Preprocessing of the input data is an additional step that is extremely important to various aspects of the deep learning training process.

The block diagram of the second stage of processing is depicted in Fig. 6. This stage consists of three parallel SdAs designed to integrate spatial and temporal context to improve decision-making. These SdAs are implemented with varying window sizes to effectively perform a multi-time-scale analysis of the signal and map event labels onto a single composite epoch label vector. A first SdA, referred to as an SPSW-SdA, is responsible for mapping labels into one of two classes: epileptiform and non-epileptiform. A second SdA, EYEM-SdA, maps labels onto the background (BCKG) and eye movement (EYEM) classes. A third SdA, 6W-SdA, maps labels to any one of the six possible classes. The first two SdAs use a relatively short window context because SPSW and EYEM are localized events and can only be detected when we have adequate temporal resolution.

Training of these three SdA networks is done in two steps: pre-training and fine-tuning. Denoising autoencoders are stacked to form a deep network. The unsupervised pre-training of such an architecture is done one layer at a time. Each layer is trained as a denoising autoencoder by minimizing the error in reconstructing its input (which is the output code of the previous layer). Once the first $k$ layers are trained, we can train the *k+1* layer because we can now compute the code or latent representation from the layer below.

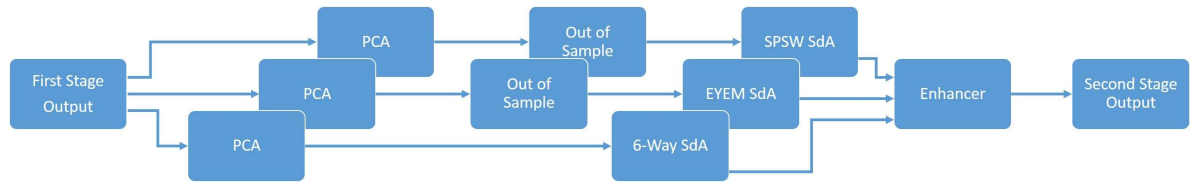

Fig. 6. An overview of the second pass of processing

Once all layers are pre-trained, the network goes through a second stage of training called fine-tuning. Here we consider supervised fine-tuning where we want to minimize prediction error on a supervised task. For this, we first add a logistic regression layer on top of the network. We then train the entire network as we would train a multilayer perceptron. At this point, we only consider the encoding parts of each auto-encoder. This stage is supervised, since now we use the target class during training (Bengio et. al., 2007; Hinton et al., 2006).

Additionally, Fig. 6 shows that input data to deep learning networks is preprocessed using a global principal components analysis (PCA) to reduce the dimensionality before applying it to these SdAs (van der Maaten, 2009). PCA is applied to each individual epoch by concatenating each channel output into a supervector and then reducing its dimensionality before it is input into SdA. For rare and localized events, which are in this case SPSW and EYEM, we use an out-of-sample technique to increase the number of training samples (van der Maaten, 2009).

Finally, using a block called an enhancer (Vincent et al., 2010), the outputs of these three SdAs are then combined to obtain the final decision. To add the three outputs together, we initialize our final probability output with the output of the 6-way classifier. For each epoch, if the other two classifiers detect epileptiform or eye movement and the 6-way classifier was not in agreement with this, we update the output probability based on the output of 2-way classifiers. The overall result of the second stage is a probability vector of dimension six containing a likelihood that each label could have occurred in the epoch. It should also be noted that the outputs of these SdAs are in the form of probability vectors. A soft decision paradigm is used rather than hard decisions because this output is smoothed in the third stage of processing.

## 2.5. Third Pass: Statistical Language Modeling

Neurologists generally impose certain restrictions on events when interpreting an EEG. For example, PLEDs and GPEDs don't happen in the same session. None of the previous stages of processing address this problem. Even the output of the second stage accounts mostly for channel context and is not extremely effective at modelling long-term temporal context. The third pass of processing addresses this issue and improves the overall detection performance by using a finite state machine based on a statistical language model.

As is shown in Fig. 1, the third stage of postprocessing is designed to impose some contextual restrictions on the output of the second stage. These contextual relationships involve long-term behavior of the signal and are learned in a data-driven fashion. This approach is also borrowed from speech recognition where a probabilistic grammar is used that combines the left and right contexts with the labels (Levinson, 2005). This is done using a finite state machine that imposes specific syntactic constraints.

In this study, a bigram probabilistic language model that provides the probability of transiting from one type of epoch to another (e.g. PLED to PLED) is prepared using the training dataset and also in consultation with neurologists in Temple Hospital University. The bigram probabilities for each of the six classes are shown in Table 1, which models all possible transitions from one label to the next. The remaining columns alternate between the class label being transitioned to and its associated probability. The probabilities in this table are optimized on a training database that is a subset of TUH-EEG. For example, since PLEDs are long-term events, the probability of transitioning from one PLED to the next is high – approximately 0.9. However, since spikes that occur in groups are PLEDs or GPEDs, and not SPSWs, the probability of transitioning from a PLED to SPSW is 0.0. Therefore, these transition probabilities emulate the contextual knowledge used by neurologists.

**Table 1. A bigram probabilistic language model for the third pass of processing which models all possible transitions from one of the six classes to the next.**

| i | j | P(i,j) | j | P(i,j) | j | P(i,j) | j | P(i,j) | j | P(i,j) | j | P(i,j) |
|------|------|------|------|------|------|------|------|------|------|------|------|------|
| SPSW | SPSW | 0.40 | PLED | 0.00 | GPED | 0.00 | EYEM | 0.10 | ARTF | 0.20 | BCKG | 0.30 |
| PLED | SPSW | 0.00 | PLED | 0.90 | GPED | 0.00 | EYEM | 0.00 | ARTF | 0.05 | BCKG | 0.05 |
| GPED | SPSW | 0.00 | PLED | 0.00 | GPED | 0.60 | EYEM | 0.00 | ARTF | 0.20 | BCKG | 0.20 |
| EYEM | SPSW | 0.10 | PLED | 0.00 | GPED | 0.00 | EYEM | 0.40 | ARTF | 0.10 | BCKG | 0.40 |
| ARTF | SPSW | 0.23 | PLED | 0.05 | GPED | 0.05 | EYEM | 0.23 | ARTF | 0.23 | BCKG | 0.23 |
| BCKG | SPSW | 0.33 | PLED | 0.05 | GPED | 0.05 | EYEM | 0.23 | ARTF | 0.13 | BCKG | 0.23 |

After compiling the probability table, a long window is centered on each epoch and the posterior probability vector for that epoch is updated by considering left and right context as a prior (essentially predicting the current epoch from its left and right context). A Bayesian framework is used to update the probabilities of this grammar for a single iteration of the algorithm:

$$P_{gprior} = \frac{\sum_{i=1}^{L} P_i + \epsilon_{prior} M}{L+M}, \tag{13}$$

$$RPP(k) = \frac{\beta_R \sum_{i=1}^{N} \exp(-i\lambda) P_{k+i} + \alpha P_{gprior}}{1+\alpha}, \tag{14}$$

$$LPP(k) = \frac{\beta_L \sum_{i=1}^{N} \exp(-i\lambda) P_{k-i} + \alpha P_{gprior}}{1+\alpha}, \tag{15}$$

$$P_{C_k|LR} = \beta_C P_{C_k} (\sum_{i=1}^{k} \sum_{j=1}^{k} LPP(i) RPP(j) Prob(i,k) Prob(k,j))^{\frac{\gamma}{n}}. \tag{16}$$

In these equations, $k = 1, 2... K$ where $K$ is the total number of classes (in this study $K = 6$), $L$ is number of epochs in a file, $\epsilon_{prior}$ is the prior probability for an epoch (a vector of length $K$) and $M$ is the weight. $LPP$ and $RPP$ are left and right context probabilities respectively. $\lambda$ is the decaying weight for window, $\alpha$ is the weight associated with $P_{gprior}$ and $\beta_R$ and $\beta_L$ are normalization factors. $P_{C_k}$ is the prior probability, $P_{C_k|LR}$ is the posterior probability of epoch $C$ for class $k$ given the left and right contexts, $y$ is the grammar weight, $n$ is the iteration number (starting from $1$) and $\beta_C$ is the normalization factor. $Prob(i,j)$ is a representation of the probability table shown in Table 1. The algorithm iterates until the label assignments, which are decoded based on a probability vector, converge. The output of this stage is the final output and what was used in the evaluations described in Section 3.

## 3. RESULTS

In this section, we present results on a series of experiments designed to optimize and evaluate each stage of processing. We used s subset of TUH-EEG for these experiments.

### 3.1. Data: The TUH-EEG Event Short Set

We collaborated with several neurologists and a team of undergraduate annotators (Shah et al., 2018)

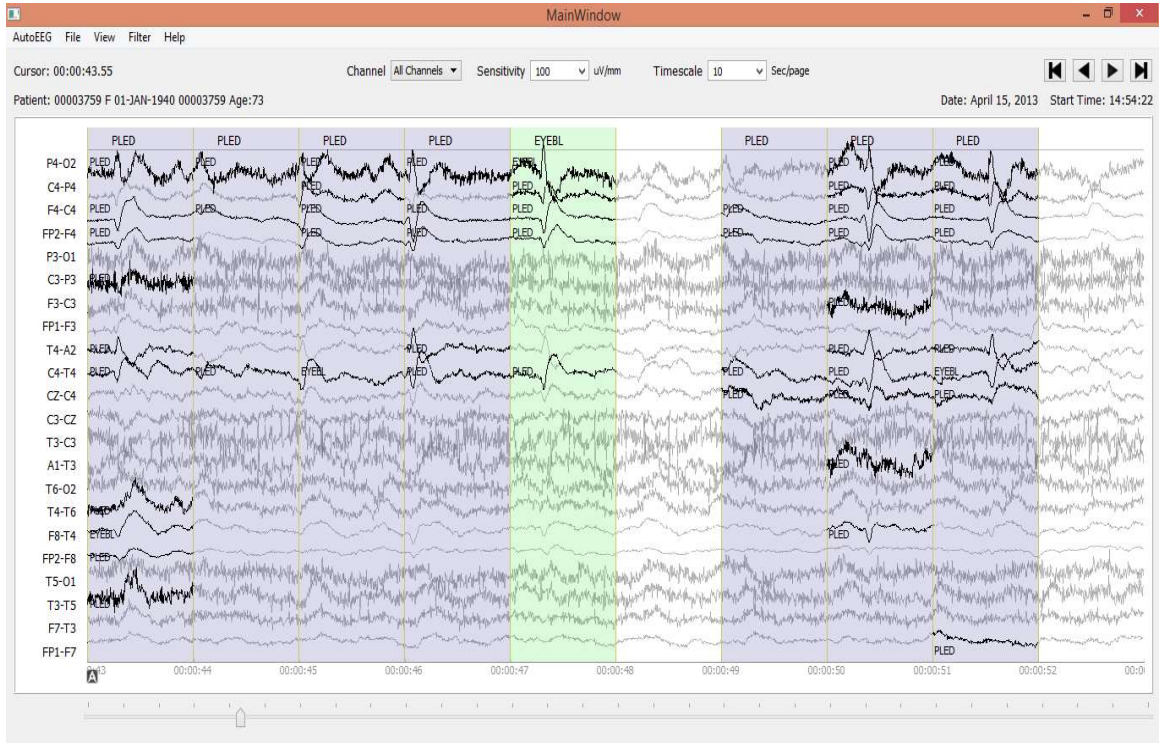

Fig. 7. An example demonstrating that the reference data is annotated on a per-channel basis.

to manually label a subset of TUH-EEG for the six types of events described in Section 2.1. The training set contains segments from 359 sessions while the evaluation set was drawn from 159 sessions. No patient appears more than once in the entire subset, which we refer to as the TUH-EEG Event Short Set (TU-EEG-ESS). Note that the annotations were created on a channel basis – the specific channels on which an event was observed were annotated. This is in contrast to many open source databases that we have observed which only mark events in time and do not annotate the specific channels on which the events occurred. In general, with EEG signals, events such as SPSW do not appear on all channels. The subset of channels on which the event appears is relevant diagnostic information. Our annotations are demonstrated in Fig. 7.

A distribution of the frequency of occurrence of the 6 types of events in the training and evaluation set is shown in Table 2. The training set was designed to provide a sufficient number of examples to train statistical models such as HMMs. Note that some classes, such as SPSW, occur much less frequently in the actual corpus than common events such as BCKG. In fact, 99% of the data is assigned to the class BCKG, so special care must be taken to build robust classifiers for the non-background classes. High performance detection of EEG events requires dealing with infrequently occurring events since much of the data is uninformative. This is often referred to as an unbalanced data problem, and it is quite common in many biomedical applications. Hence, the evaluation set was designed to contain a reasonable representation of all classes. All of EEGs in this subset were recorded using standard 10–20 system and processed using a TCP montage (Lopez et al., 2016), resulting in 22 channels of signal data per EEG.

**Table 2. An overview of the distribution of events in the subset of the TUH EEG Corpus used in our experiments**

| Event | Train | Train % (CDF) | Eval | Eval % (CDF) |
|---|---|---|---|---|
| SPSW | 645 | 0.8% ( 1%) | 567 | 1.9% ( 2%) |
| GPED | 6,184 | 7.4% ( 8%) | 1,998 | 6.8% ( 9%) |
| PLED | 11,254 | 13.4% (22%) | 4,677 | 15.9% (25%) |
| EYEM | 1,170 | 1.4% (23%) | 329 | 1.1% (26%) |
| ARTF | 11,053 | 13.2% (36%) | 2,204 | 7.5% (33%) |
| BCKG | 53,726 | 63.9% (100%) | 19,646 | 66.8% (100%) |
| Total: | 84,032 | 100.0% (100%) | 29,421 | 100.0% (100%) |

**Table 5. The *6*-way classification results for the first pass of processing**

| Event | ARTF | BCKG | EYEM | GPED | PLED | SPSW |
|-------|------|------|------|------|------|------|
| ARTF  | 41.24 | 45.19 | 2.18 | 3.81 | 2.77 | 4.81 |
| BCKG  | 7.02 | 71.93 | 2.59 | 7.37 | 2.28 | 8.81 |
| EYEM  | 2.13 | 0.61 | 82.37 | 2.13 | 8.51 | 4.26 |
| GPED  | 7.46 | 4.85 | 2.39 | 53.32 | 20.42 | 11.55 |
| PLED  | 0.70 | 1.85 | 4.70 | 17.62 | 54.80 | 20.32 |
| SPSW  | 4.41 | 8.29 | 9.17 | 33.33 | 4.59 | 40.21 |

**Table 4. The *4*-way classification results for the first pass of processing**

| Event | BCKG | SPSW | GPED | PLED |
|-------|------|------|------|------|
| BCKG  | 82.30 | 8.35 | 6.94 | 2.42 |
| SPSW  | 21.87 | 40.21 | 33.33 | 4.59 |
| GPED  | 14.71 | 11.55 | 53.32 | 20.42 |
| PLED  | 7.26 | 20.32 | 17.62 | 54.80 |

**Table 3. The *2*-way classification results for the first pass of processing**

| Event | TARG | BCKG |
|-------|------|------|
| TARG  | 86.92 | 13.08 |
| BCKG  | 18.20 | 81.80 |

## 3.2.    Preprocessing: Feature Extraction

Features from each epoch are identified using a feature extraction technique explained in Section 2.2. Neurologists review EEGs in *10* sec windows. Pattern recognition systems often subdivide the signal into small segments during which the signal can be considered quasi-stationary. HMM systems need further subdivision so that there are enough observations to allow the system to develop a strong sense of the correct choice. A simple set of preliminary experiments determined that a reasonable tradeoff between computational complexity and performance was to split the *10* sec window into *1* sec epochs, and to further subdivide these into *0.1* sec frames. Hence, features were computed every *0.1* sec using a *0.2* sec overlapping analysis window. These parameters were optimized experimentally in a previous study (Harati et al., 2015).

We have also previously shown that the use of a novel differential energy feature improved performance for absolute features, but that benefit diminishes as first and second-order derivatives are included. We have shown there is benefit to using derivatives and there is a small advantage to using frequency domain energy. The output of the feature extraction system is *22* channels of data, where in each channel, a feature vector of dimension *26* corresponds to every *0.1* secs.

## 3.3.    First Pass: Sequential Decoding Using Hidden Markov Models

A *6*-way classification experiment was conducted using the models described in Fig. 4. Each state uses *8* Gaussian mixture components and a diagonal covariance assumption (drawing on our experience with speech recognition systems and balancing dimensionality of the models with the size of the training data).

Models were trained using all events on all channels resulting in what we refer to as channel independent models. Channel dependent models have not proven to provide a boost in performance and add considerable complexity to the system.

The results for the first pass of processing are shown in Table 5. A more informative performance analysis can be constructed by collapsing the three background classes into one category. We refer to this second evaluation paradigm as a *4*-way classification task: SPSW, GPED, PLED and BACKG. The latter class contains an enumeration of the three background classes. The *4*-way classification results for the first pass of processing are presented in Table 4. Finally, in order that we can produce a DET curve (Martin et al., 1997) we also report a *2*-way classification result in which we collapse the data into a target class (TARG) and a background class (BCKG). The *2*-way classification results for the first pass of processing are presented in Table 3. Note that the classification results for all these tables are measured by counting each epoch for each channel as an independent event. We refer to this as forced-choice event-based scoring because every epoch for every channel is assigned a score based on its class label.

### 3.4.   Second Pass: Temporal and Spatial Context Analysis Based on Deep Learning

The output of the first stage of processing is a vector of six scores, or likelihoods, for each channel at each epoch. Therefore, if we have *22* channels and six classes we will have a vector of dimension *6 x 22 = 132* scores for each epoch. This *132*-dimension epoch vector is computed without considering similar vectors from epochs adjacent in time. Information available from other channels within the same epoch is referred to as "spatial" context since each channel corresponds to a specific electrode location on the skull.

Information available from other epochs is referred to as "temporal" context. The goal of this level of processing is to integrate spatial and temporal context to improve decision-making.

To integrate context, the input to the second pass deep learning system is a vector of dimension *6 x 22* x *window length*, where we aggregate *132*-dimension vectors in time. If we consider a *41*-second window, then we will have a *5,412*-dimension input to the second pass of processing. This input dimensionality is high even though we have a considerable amount of manually labeled training. To deal with this problem we follow a standard approach of using Principal Components Analysis (PCA) (Fukunaga, 1983) before every SdA. The output of the PCA is a vector of dimension *13* for SdA detectors that look for SPSW and EYEM and *20* for *6*-way SdA classifier.

Further, since we do not have enough SPSW and EYEM events in the training dataset, we must use an out-of-sample technique (van der Maaten, 2009) to train the SdA. Three consecutive outputs are averaged, so the output is further reduced from *3 x 13* to just *13*, using a sliding window approach to averaging. Therefore, the input for SPSW SdA and EYEM SdA decreases to *13 x window length* and *20 x window length* for *6*-way SdA.

We used an open source toolkit, Theano (Bastien et al., 2012; Bergstra et al., 2010), to implement the SdAs. The parameters of the models are optimized to minimize the average reconstruction error using a cross-entropy loss function. In the optimization process, a variant of stochastic gradient descent is used, referred to as minibatches. Minibatch stochastic gradient descent works identically to stochastic gradient descent, except that we use more than one training example to make each estimate of the gradient. This technique reduces variance in the estimate of the gradient, and often makes better use of the hierarchical memory organization in modern computers.

SPSW SdA uses a window length of *3* which means it has *39* inputs and *2* outputs. It has three hidden layers with corruption levels of *0.3* for each layer. The number of nodes per layer are: first layer = *100*, second layer = *100*, third layer = *100*. The parameters for pre-training are: learning rate = *0.5*, number of

**Table 8. The *6*-way classification results for the second pass of processing**

| Event | ARTF | BCKG | EYEM | GPED | PLED | SPSW |
|-------|------|------|------|------|------|------|
| ARTF | 27.49 | 61.73 | 7.28 | 0.00 | 1.08 | 2.43 |
| BCKG | 7.00 | 82.03 | 5.79 | 0.97 | 0.36 | 3.86 |
| EYEM | 4.21 | 16.84 | 77.89 | 0.00 | 0.00 | 1.05 |
| GPED | 0.60 | 14.69 | 0.00 | 59.96 | 10.26 | 14.49 |
| PLED | 1.40 | 22.65 | 0.80 | 13.83 | 52.30 | 9.02 |
| SPSW | 7.69 | 35.90 | 2.56 | 28.21 | 0.00 | 25.64 |

**Table 7. The *4*-way classification results for the second pass of processing**

| Event | BCKG | SPSW | GPED | PLED |
|-------|------|------|------|------|
| BCKG | 95.60 | 3.24 | 0.62 | 0.54 |
| SPSW | 46.15 | 25.64 | 28.21 | 0.00 |
| GPED | 15.29 | 14.49 | 59.96 | 10.26 |
| PLED | 24.85 | 9.02 | 13.83 | 52.30 |

**Table 6. The *2*-way classification results for the second pass of processing**

| Event | TARG | BCKG |
|-------|------|------|
| TARG | 78.94 | 21.06 |
| BCKG | 4.40 | 95.60 |

epochs = *200*, batch size = *300*. The parameters for fine-tuning are: learning rate = *0.2*, number of epochs = *800* and batch size = *100*.

EYEM SdA uses a window length of *3* which means it has *39* inputs and *2* outputs. It has three hidden layers with corruption levels of *0.3* for each layer. The number of nodes per layer are: first layer = *100*, second layer = *100*, third layer = *100*. The parameters for pre-training are: learning rate = *0.5*, number of epochs = *200*, batch size = *300*. The parameters for fine-tuning are: learning rate = *0.2*, number of epochs = *100* and batch size = *100*.

Six-way SdA uses a window length of *41* which means it has *820* inputs and *6* outputs. It has three hidden layers with corruption levels of *0.3* for each layer. The number of nodes per layer are: first layer = *800*, second layer = *500*, third layer = *300*. The parameters for pre-training are: learning rate = *0.5*, number of epochs = *150* and batch size = *300*. The parameters for fine-tuning are: learning rate = *0.1*, number of epochs = *300* and batch size = *100*.

The *6*-way, *4*-way and *2*-way classification results for the second stage of processing are presented in Table 8, Table 7, Table 6 respectively. Note that unlike the tables for the first pass of processing, the classification results in each of these tables are measured once per epoch – they are not per-channel results. We refer to these results as epoch-based.

### 3.5.   Third Pass: Statistical Language Modeling

The output of the second stage of processing is a vector of six scores, or likelihoods, per epoch. This serves as the input for the third stage of processing. The optimized parameters for the third pass of processing are: prior probability for an epoch, $\epsilon_{prior}$, is *0.1*; the weight, *M*, is *1*; the decaying weight, $\lambda$, is *0.2*; the weight associated with $P_{gprior}$, $\alpha$, is *0.1*; the grammar weight, *y*, is *1*; the number of iterations, *n*, is *20,* and the window length to calculate the left and right prior probabilities is *10*.

The *6*-way, *4*-way and *2*-way classification results are presented in Table 11, Table 9 and Table 10 respectively. Note that these results are also epoch-based.

**Table 11. The *6*-way classification results for the third pass of processing**

| Event | ARTF | BCKG | EYEM | GPED | PLED | SPSW |
|-------|------|------|------|------|------|------|
| ARTF | 14.04 | 72.98 | 10.18 | 0.00 | 0.00 | 2.81 |
| BCKG | 3.42 | 81.40 | 8.93 | 0.30 | 0.00 | 5.95 |
| EYEM | 2.30 | 17.24 | 79.31 | 0.00 | 0.00 | 1.15 |
| GPED | 0.30 | 3.65 | 0.00 | 65.05 | 13.37 | 17.63 |
| PLED | 0.00 | 10.76 | 0.49 | 9.78 | 65.28 | 13.69 |
| SPSW | 10.00 | 33.33 | 13.33 | 10.00 | 0.00 | 33.33 |

**Table 9. The *4*-way classification results for the third pass of processing**

| Event | BCKG | SPSW | GPED | PLED |
|-------|------|------|------|------|
| BCKG | 95.11 | 4.69 | 0.19 | 0.00 |
| SPSW | 56.67 | 33.33 | 10.00 | 0.00 |
| GPED | 3.95 | 17.63 | 65.05 | 13.37 |
| PLED | 11.25 | 13.69 | 9.78 | 65.28 |

**Table 10. The *2*-way classification results for the third pass of processing**

| Event | TARG | BCKG |
|-------|------|------|
| TARG | 90.10 | 9.90 |
| BCKG | 4.89 | 95.11 |

## 4. DISCUSSION

The *6*-way classification task can be structured into several subtasks. Of course, due to the high probability of the signal being background, the system is heavily biased towards choosing the background model. Therefore, in Table 4 we see that performance on BACKG is fairly high. Not surprisingly, BCKG is most often confused with SPSW. SPSW events are short in duration and there are many transient events in BCKG that resemble an SPSW event. This is one reason we added ARTF and EYEM models, so that we can reduce the confusions of all classes with the short impulsive SPSW events. As we annotate background data in more detail, and identify more commonly occurring artifacts, we can expand on our ability to model BCKG events explicitly.

GPEDs are, not surprisingly, most often confused with PLED events. Both events have a longer duration than SPSWs and artifacts. From Table 4, we see that performance on these two classes is generally high. The main difference between GPED and PLED is duration, so we designed the postprocessing to learn this as a discriminator. For example, in the second pass of processing, we implemented a window duration of 41 seconds so that the SdA system would be exposed to long-term temporal context. We also designed three separate SdA networks to differentiate between short-term and long-term context. In Table 7 we see that the performance of GPEDs and PLEDs improves with the second pass of postprocessing. More significantly, the confusions between GPEDs and PLEDs also decreased. Note also that in Table 7 performance of BCKG increased significantly. Confusions with GPEDs and PLEDs dropped dramatically to below 1%.

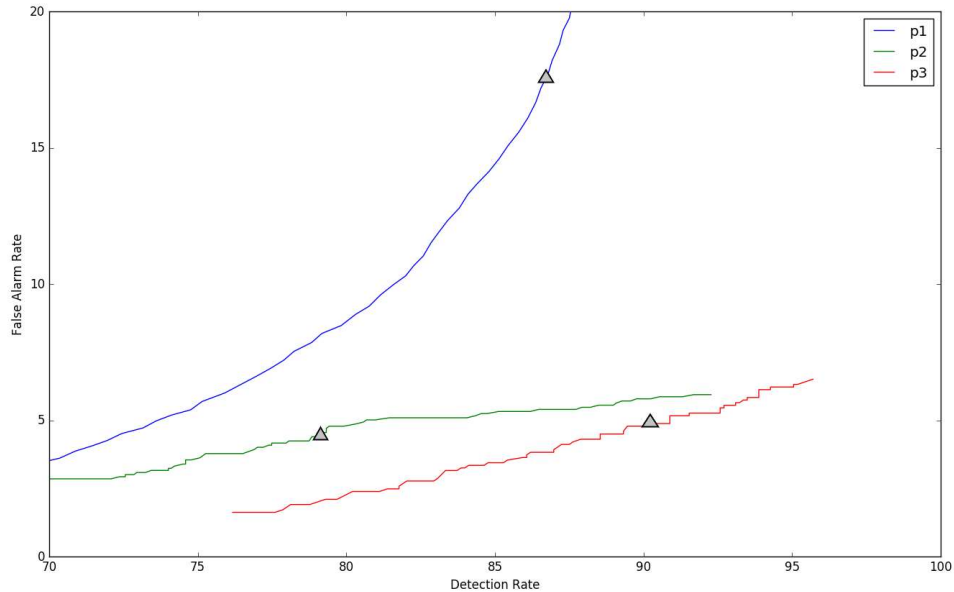

Fig. 8. DET curves are shown for each pass of processing. The "zero penalty" operating point is also shown since this was used in Table 5 – Table 10.

While performance across the board increased, performance for SPSW dropped by adding the second pass of postprocessing. This is a reflection on the imbalance of the data. Less than one percent of data is annotated as SPSWs, while we have ten times more training samples for GPEDs and PLEDs. Note that we used an out-of-sample technique to increase the number of training samples for SPSWs, but even this technique could not solve the problem of a lack of annotated SPSW data. By comparing Table 5 to Table 8 we saw a similar behavior with the EYEM class because there is also fewer EYEM events.

A summary of the results for different stages of processing is shown in Table 12. The overall performance of the multi-pass hybrid HMM/deep learning classification system is promising: more than 90% sensitivity and less than 5% specificity.

Because the false alarm rate in these types of applications varies significantly with sensitivity, it is important to exam performance using a DET curve. A DET curve for the first, second and third stage of processing is given in Fig. 8. Note that the tables previously presented use the unprocessed likelihoods output from the system. They essentially correspond to the point on the DET curve where a penalty of 0 is applied. This operating point is identified on each of the curves in Fig. 8. We see that the raw likelihoods of the system correspond to different operating points in the DET curve space. From Fig. 8 it is readily apparent that postprocessing significantly improves our ability to maintain a low false alarm rate as we increase the detection rate. In virtually all cases, the trends shown in Table 5 to Table 12 hold up for the full range of the DET curve. This study demonstrates that a significant amount of contextual processing is required to achieve a specificity of 5%.

**Table 12. Specificity and sensitivity for each pass of processing**

| Pass | Sensitivity | Specificity |
|------|-------------|-------------|
| 1 (HMM) | 86.78 | 17.70 |
| 2 (SdA) | 78.93 | 4.40 |
| 3 (SLM) | 90.10 | 4.88 |

## 5.  CONCLUSION

Virtually all previous R&D efforts involving EEG, including seizure detection, have been conducted on small databases (Akareddy et al., 2014). Often these databases are not good representations of the type of data observed in clinical environments. Transient artifacts, not common in databases collected under research conditions, can significantly degrade performance. Not surprisingly, despite high accuracies presented in the research literature, the performance of commercially available systems has been lacking in clinical settings. There is still great demand for an automated system that achieves a low false alarm rate in clinical applications.

We have presented a three-pass system that can achieve high performance classifying EEG events of clinical relevance. The system uses a combination of HMMs for accurate temporal segmentation and deep learning for high performance classification. In the first pass, the signal is converted to EEG events using a hidden Markov model based system that models the temporal evolution of the signal. In the second pass, three stacked denoising autoencoders (SDAs) with different window durations are used to map event labels onto a single composite epoch label vector. We demonstrated that both temporal and spatial context analysis based on deep learning can improve the performance of sequential decoding using HMMs. In the third pass, a probabilistic grammar is applied that combines left and right context with the current label vector to produce a final decision for an epoch.

Our hybrid HMM/deep learning system delivered a sensitivity above 90% while maintaining a specificity below 5%, making automated analysis a viable option for clinicians. This framework for automatic analysis of EEGs can be applied in other classification tasks such as seizure detection or abnormal detection. There are many straightforward extensions of this system that can include more powerful deep learning networks such as Long Short-Term Memory Networks or Convolutional Neural Networks. This is the subject of our ongoing research.

This project is part of a long-term collaboration with the Department of Neurology at Temple University Hospital that has produced several valuable outputs including a large corpus (TUH-EEG), a subset of the corpus annotated for clinically relevant events (TUH-EEG-ESS), and technology to automatically interpret EEGs. In related work, we are also making the corpus searchable using multimodal queries that integrate metadata, information extracted from EEG reports and the signal event data described here (Picone et al., 2016). The resulting system can be used to retrieve many different types of cohorts and will be a valuable tool for clinical work, research and teaching.

### ACKNOWLEDGEMENTS

The primary funder of this research was the QED Proof of Concept program of the University City Science Center (Grant No. S1313). Research reported in this publication was also supported by the National Human Genome Research Institute of the National Institutes of Health under Award Number U01HG008468 and the National Science Foundation through Major Research Instrumentation Grant No. CNS-09-58854. The TUH-EEG database work was funded by (1) the Defense Advanced Research Projects Agency (DARPA) MTO under the auspices of Dr. Doug Weber through the Contract No. D13AP00065, (2) Temple University's College of Engineering and (3) Temple University's Office of the Senior Vice-Provost for Research. Finally, we are also grateful to Dr. Mercedes Jacobson, Dr. Steven Tobochnik and David Jungries of the Temple University School of Medicine for their assistance in developing the classification paradigm used in this study and for preparing the manually annotated data.